\documentclass[preprint,12pt,num]{elsarticle}

\usepackage{amssymb}
\usepackage{amsmath}
\usepackage{booktabs} 
\usepackage{subcaption}
\usepackage{longtable}
\usepackage{multirow} 
\usepackage{adjustbox} % For table fitting
\usepackage[symbol]{footmisc} % Allows footnote symbols

\journal{arXiv}

\begin{document}

\begin{frontmatter}



\title{Exploring Large Protein Language Models in Constrained Evaluation Scenarios within the FLIP Benchmark} %% Article title


\author{Manuel F. Mollon} %% Author name
\author{Joaquin Gonzalez-Rodriguez} %% Author name
\author{Alicia Lozano-Diez} %% Author name
\author{Daniel Ramos} %% Author name
\author{Doroteo T. Toledano} %% Author name

\affiliation{organization={AUDIAS},%Department and Organization
            addressline={Universidad Autonoma de Madrid (UAM)},
            country={España}}

%% Abstract
\begin{abstract}
In this study, we expand upon the FLIP benchmark—designed for evaluating protein fitness prediction models in small, specialized prediction tasks—by assessing the performance of state-of-the-art large protein language models, including ESM-2 and SaProt on the FLIP dataset. Unlike larger, more diverse benchmarks such as ProteinGym, which cover a broad spectrum of tasks, FLIP focuses on constrained settings where data availability is limited. This makes it an ideal framework to evaluate model performance in scenarios with scarce task-specific data. We investigate whether recent advances in protein language models lead to significant improvements in such settings. Our findings provide valuable insights into the performance of large-scale models in specialized protein prediction tasks.
\end{abstract}



%% Keywords

\end{frontmatter}



\section{Introduction}
\label{introduction}
%% Labels are used to cross-reference an item using \ref command.

Protein fitness prediction, the process of forecasting the functional impact of mutations on protein behavior, is a critical task in computational biology. Accurate prediction of protein fitness has significant implications for various fields, including drug design, protein engineering, and directed evolution. Directed evolution \cite{directed_evolution}, in particular, relies on the ability to predict how mutations will influence protein properties, enabling the design of proteins with desired traits. However, traditional methods of protein fitness prediction, such as early machine learning (ML)-based approaches often require large datasets, limiting their application to certain scenarios.

To address this challenge, the FLIP (Functional Landscape of Interacting Proteins) \cite{flip} benchmark was developed as a specialized dataset tailored to predict protein fitness under constrained conditions. Unlike larger datasets, such as ProteinGym, which encompass a wide array of protein-related tasks, the FLIP dataset focuses on smaller, more specific tasks. FLIP uses data splits like two vs many and low vs high, providing a unique opportunity to study model performance in scenarios with lower mutation levels during training and higher mutation levels during testing. This differs from the ProteinGym \cite{ProteinGym} approach, which involves random sampling across tasks. The FLIP setup is important for real-world applications where models must generalize well to more complex, higher-mutation scenarios. This structure makes overfitting a bigger concern in FLIP compared to random sampling scenarios, as the model is trained on lower mutations and tested on higher mutations. While this does not inherently make one approach better or worse, it is a crucial factor to consider when analyzing training and evaluation metrics in this paper, especially for models with limited data and a need for generalization across diverse protein functions.

Recent advancements in state-of-the-art models like ESM-2 \cite{esm2_esmfold} and SaProt \cite{saprot} have shown promising results in addressing the challenge of requiring large datasets by using self-supervised pretraining. These models are trained on vast amounts of unlabeled protein sequence data, enabling them to capture patterns and structural properties of proteins without needing labeled datasets. This allows them to generalize well in low-data scenarios or zero-shot tasks. These models are increasingly being explored for protein fitness prediction tasks, where their ability to generalize from limited training data is highly valuable. In this study, we expand on the FLIP baseline by testing the performance of these recent high-performance models within the context of the FLIP benchmark, aiming to understand how they perform in small, task-specific datasets.

\section{Dataset}

Within the FLIP benchmark, several activities provide the foundational tasks for protein fitness prediction. Among them, three key activities are GB1, Meltome, and AAV. These three tasks used in this paper and summarized in this section represent distinct challenges in protein fitness prediction, each contributing to our understanding of how different protein features influence function under mutation.
For evaluation, we selected a subset of the splits used in the FLIP benchmark. The selected splits and their sample distributions across train and test sets are shown in Table~\ref{tab:sample_distribution}. For detailed information on each split, please refer to the FLIP paper \cite{flip}.

These activities within the FLIP benchmark provide a rich and diverse set of tasks that require accurate protein fitness predictions. The tasks vary in their focus, from stability and melting temperature predictions to the functional efficiency of viral vectors, making the FLIP benchmark ideal for testing protein language models in small, task-specific scenarios.

\subsection{AAV}
The AAV task focuses on predicting the efficiency of adenovirus-associated virus (AAV) vectors used in gene therapy. The task involves predicting how mutations in the AAV capsid affect its ability to deliver genetic material to target cells. This is an important application in gene therapy, as optimizing the viral vector’s efficiency can lead to better therapeutic outcomes. The AAV dataset consists of mutations that either enhance or reduce the viral vector’s transduction efficiency, and the challenge is to predict these effects based on sequence data.

\subsection{Meltome (Thermostability)}
The Meltome activity involves predicting the melting temperature (\(T_m\)) of proteins, which serves as a proxy for protein stability. The melting temperature is an important characteristic that reflects the protein’s ability to maintain its functional conformation under different conditions. The Meltome dataset includes a wide range of mutations across various proteins, and the task is to predict how these mutations will shift the \(T_m\) value, indicating changes in protein stability. This task is particularly relevant for protein design, where maintaining stability is often a key goal.

\subsection{GB1}
The GB1 task focuses on the prediction of the stability of a single-domain antibody, GB1. This is a critical task for understanding antibody engineering. The dataset for GB1 contains mutations that are known to either stabilize or destabilize the protein, and the challenge lies in determining how these mutations influence the overall structural integrity and binding affinity.

\begin{table}[!ht]
\centering
\begin{tabular}{llrrr}
\toprule
\textbf{Landscape} & \textbf{Split} & \textbf{Total} & \textbf{Train} & \textbf{Test} \\
\midrule
\multirow{3}{*}{AAV} & 2-vs-rest & 82,583 & 31,807 & 50,776 \\
                     & 7-vs-rest & 82,583 & 70,002 & 12,581 \\
                     & low-vs-high & 82,583 & 47,546 & 35,037 \\
\midrule
\multirow{2}{*}{Meltome} & Human & 10,093 & 8,148 & 1,945 \\
                                 & Human-cell & 7,156 & 5,792 & 1,366 \\
\midrule
\multirow{3}{*}{GB1} & 2-vs-rest & 8,733 & 427 & 8,306 \\
                     & 3-vs-rest & 8,733 & 2,968 & 5,765 \\
                     & low-vs-high & 8,733 & 5,089 & 3,644 \\
                     & sampled & 8,733 & 6,961 & 1,772 \\
\bottomrule
\end{tabular}
\caption{Sample distribution across landscapes and splits}
\label{tab:sample_distribution}
\end{table}

\section{Evaluated Models}

In this study, we extend the evaluation of the FLIP benchmark to include additional state-of-the-art models and enhance training procedures to improve performance. Below, we summarize each model, including modifications made to ensure a comprehensive and fair assessment of model capabilities on the dataset.

\subsection{ESM-1v \cite{esm}}
The ESM 1v model was previously used in the FLIP baseline; however, in this study, we make several improvements. The training process was extended by increasing the number of epochs and employing a learning rate (LR) scheduler to dynamically adjust the learning rate based on model performance, with early stopping applied via a patience parameter to prevent overfitting. These modifications aim to optimize the performance of the model in protein fitness prediction tasks.

\subsection{ESM-2}

We evaluate three configurations of the ESM-2 model: a 6, 33, and 48 layer variant. Table \ref{tab:esm_models} provides details regarding the parameter count for each model. The 6-layer and 48-layer models were selected to assess the smallest and largest ESM-2 configurations, allowing us to analyze how model size impacts learning capabilities and generalization power. Additionally, we included the 33-layer model as an intermediate option to compare the performance of ESM-2 with its predecessor, ESM-1v, since both have approximately the same number of parameters.

\begin{table}[!ht]
\centering
\begin{tabular}{lllll}
\toprule
\textbf{Model Name}       & \textbf{Layers} & \textbf{Params} & \textbf{Embedding Dim} & \textbf{Alias} \\ 
\midrule
esm2\_t48\_15B\_UR50D    & 48              & 15B             & 5120                   & esm2\_48            \\ 
esm2\_t33\_650M\_UR50D   & 33              & 650M            & 1280                   & esm2\_33            \\ 
esm2\_t6\_8M\_UR50D      & 6               & 8M              & 320                    & esm2\_6             \\ 
\bottomrule
\end{tabular}
\caption{Summary of ESM models with their respective configurations.}
\label{tab:esm_models}
\end{table}

ESM-2, one of the latest iteration in the ESM model family, introduces architectural improvements aimed at enhancing depth and representational capacity, thereby improving performance on protein-related tasks. By benchmarking these ESM-2 configurations, we seek to evaluate the impact of model depth on fitness prediction tasks, particularly in terms of generalization, overfitting, and robustness across different dataset splits. This analysis not only updates benchmark results to reflect the capabilities of the latest ESM models, but also provides insights into the trade-offs between model complexity and performance, offering valuable guidance for selecting the most appropriate model for specific tasks in protein engineering.

\subsection{SaProt}
SaProt, a structure-aware model, is designed to incorporate structural information into the prediction of protein fitness. Specifically, it focuses on sequence patterns associated with both functional and structural protein attributes. By including SaProt in the benchmark, we aim to evaluate how well models can leverage structural information to enhance their predictions.

The pipeline used to implement SaProt consists of several steps. First, protein structure is predicted using ESMFold \cite{esm2_esmfold}, an efficient alternative to AlphaFold2 (AF2). ESMFold generates predicted protein structures as atomic coordinate files in the Protein Data Bank format (``.pdb''
), which are then processed using FoldSeek \cite{foldseek} to convert these files into structure characters corresponding to each amino acid. This conversion allows the generation of a structure-aware sequence, enhancing the model's understanding of the protein’s function in the context of its 3D conformation.

In our experiments, we use pLTTD (predicted Local True Template Distance) values as a measure of structural accuracy. In the first experiment, we use unmasked sequences, including all amino acids in their native forms. In a second experiment, we masked amino acids with a pLTTD lower than seventy. This allows us to assess the effect of masking less confident structural predictions on the model's ability to use structural information effectively. This approach provides insights into how SaProt’s performance is impacted by varying levels of structural accuracy, highlighting the importance of structural quality for enhancing predictions in protein fitness tasks.

\subsection{Considerations for a Fair Benchmark}

To ensure fairness in model comparison, a strict distinction was maintained between training, validation, and test sets across all models. These splits were predefined and remain consistent across all evaluations, ensuring that no model sees the same sequence in both training and testing phases. Sequence-aware models were allowed to leverage the sequences to inform structural prediction, but no explicit structural information about the proteins was added to the model inputs beyond the sequences themselves. This consistency is critical for fair assessment, as it ensures that all models are evaluated on an equal footing, with no additional prior information about the proteins influencing the predictions.

\section{Evaluation Framework}

\subsection{Training Details}

\begin{figure}[!ht]
    \centering
    % Example for one split
    \includegraphics[width=0.5\textwidth]{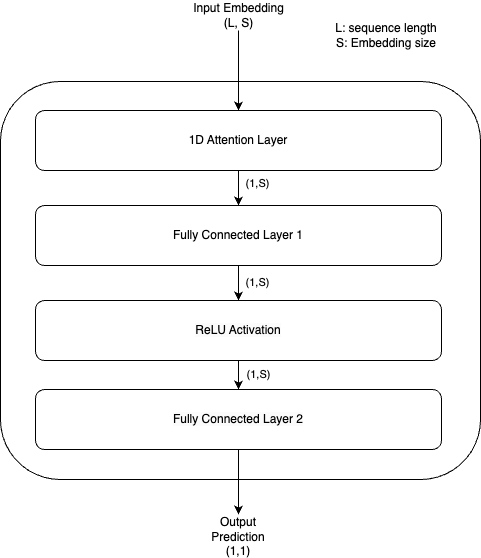}
    \caption{Model that process embeddings}
    \label{fig:small_model}
\end{figure}

\begin{figure}[!ht]
    \centering
    % Example for one split
    \includegraphics[width=0.99\textwidth]{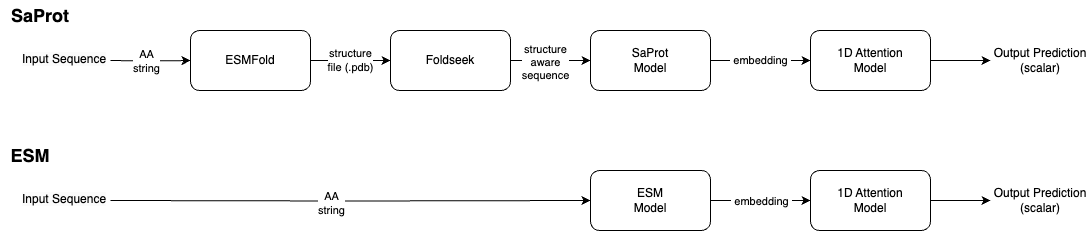}
    \caption{Pipeline implementation for SaProt and ESM}
    \label{fig:pipeline}
\end{figure}

For training, the models were used as embedding extractors. These resulting embeddings were then used to train a lightweight model designed to process input tensors, where each sequence is represented by a set of embeddings. This model was borrowed from the one used in the FLIP paper code. The `Attention1d` layer computes attention weights over the sequence dimension, generating a weighted representation that effectively summarizes the entire sequence into a single tensor. This summarized representation is passed through a fully connected layer, followed by a ReLU activation to introduce non-linearity. Finally, the transformed embedding is fed into another linear layer, which outputs a scalar value used for the downstream classification task. For details on the model refer to Figure \ref{fig:small_model} and for pipeline implementation refer to Figure \ref{fig:pipeline}.

% Loop through each split and metric

The lightweight models were fine-tuned for a maximum of 500 epochs, starting with an initial learning rate of 0.001. A learning rate scheduler was used to dynamically adjust the learning rate based on the validation loss, and early stopping with a patience of 20 epochs was implemented to prevent overfitting on the validation set. These settings were carefully chosen to promote consistent and stable convergence across all models, enabling a fair comparison of their performance under identical fine-tuning conditions. To evaluate consistency and variability in convergence, each model was trained over 10 independent experiments, using the same set of 10 random seeds for reproducibility.

However, it is important to highlight that these measures do not completely prevent the model from overfitting to the specific distribution of mutations present in the training set. Such overfitting can negatively impact the model’s ability to generalize to test samples with higher numbers of mutations. This limitation can be evaluated by comparing the train-to-test performance ratios within the sampled split to those observed in other splits, providing insight into the extent of overfitting and generalization issues.

\subsection{Evaluation Metrics: Mean Squared Error and Spearman Rank Correlation}

Mean Squared Error (MSE) and Spearman rank correlation were chosen as evaluation metrics to assess both predictive accuracy and rank reliability. MSE serves as a standard metric to gauge the absolute error in predictions, highlighting the model's overall accuracy by penalizing large errors. Meanwhile, Spearman rank correlation, a non-parametric measure of rank correlation, assesses how well the predictions preserve the order of protein fitness scores.

In protein fitness landscape applications, Spearman rank correlation ($\rho$) is particularly valuable because it evaluates the model's ability to capture the relative ordering of fitness levels rather than absolute values, which are often less critical. This metric is widely used in fitness landscape modeling as it can effectively reflect whether a model correctly ranks sequences according to their relative fitness. High Spearman correlation values indicate that the model accurately captures the underlying fitness landscape, enabling researchers to predict protein stability and activity across sequence variations effectively. Given the typical challenges in protein modeling, including small sample sizes and noisy fitness measurements, the use of Spearman rank correlation provides a robust assessment of a model's generalization and consistency in ranking protein fitness across variations.

\subsection{Implementation}

The FLIP repository was used as the starting point for training and evaluating the ESM models. However, several modifications were made to adapt it to our specific requirements. These included adjusting the training settings to incorporate early stopping and a learning rate scheduler, as previously mentioned, as well as adding functionality to calculate embeddings for the newer ESM models. For SaProt, we utilized the data splits provided by the FLIP repository but developed new code for structure prediction using ESMFold and FoldSeek, along with implementations for training, evaluating, and metric summarization. The ESMFold and SaProt repository were used for structure and embedding prediction, respectively.

\subsection{Computational Resources}

For running experiments and accessing computational resources, we utilized the CCC (Centro de Computación Científica - UAM). Specifically, we performed our tasks using four NVIDIA A100 80GB PCIe GPUs. For more information, please refer to the CCC website\footnote[1]{https://www.ccc.uam.es}.

\section{Results}

In this section, we present the results of each model's performance on the protein fitness prediction tasks, evaluated using mean squared error (MSE) and Spearman's rank correlation coefficient. The results shown represent the median values of ten experiments run per split, as the median is more robust to outliers compared to the average. Detailed visualizations of these metrics for each split can be found in the Appendix, where violin plots illustrate the distribution of MSE and Spearman's rank coefficients across splits and experiments.

Tables \ref{tab:meltome_human} through \ref{tab:aav_low_vs_high} present the evaluation results across different models and splits. The `\textit{esm1v}' results were not computed by us; rather, they were directly taken from the supplementary material of the original FLIP paper \cite{flip}. The results highlight the variations in predictive performance between models, providing valuable insights into the strengths and weaknesses of each architecture for protein fitness prediction. A general trend observed is a slight improvement in both performance and generalization ability with the newer ESM-2 models compared to their ESM1 counterparts of the same size, along with a more significant improvement in the larger ESM-2 models. Overall, SaProt and the ESM-2 model with 33 layers achieve the most consistent results. However, caution is needed, as they are more prone to overfitting than smaller or simpler models.

\subsection{Meltome}

\begin{table}[ht]
\centering
\caption{Results for Meltome (human)}
\begin{tabular}{lcc cc}
\toprule
Model & \multicolumn{2}{c}{$\rho$} & \multicolumn{2}{c}{MSE} \\
\cmidrule(lr){2-3} \cmidrule(lr){4-5}
                            & Train             & Test              & Train             & Test              \\
\midrule
\textit{esm1v}\textsuperscript{\cite{flip}}                & 0.770             & 0.690             & 8.96             & 12.59             \\
esm1v (6 Layers)            & 0.790             & 0.690             & 8.51             & 12.49             \\
esm2 (6 Layers)             & 0.715             & 0.680             & 11.0            & 14.01             \\
esm2 (33 Layers)            & 0.780             & 0.700             & 8.56             & 12.35             \\
esm2 (48 Layers)            & \textbf{0.955}    & 0.660             & \textbf{1.87}    & 15.79             \\
SaProt (ESMFold unmasked) \footnote[2]  & 0.777             & \textbf{0.718}    & 8.76            & \textbf{11.47}  \\
SaProt (ESMFold masked) \footnote[2]   & 0.779             & 0.717             & 8.70            & 11.76           \\
\bottomrule
\end{tabular}
\label{tab:meltome_human}
\end{table}

\footnotetext[2]{Sequences longer than 2700 AA (122 samples) were masked completely due to computational constraints.}

\begin{table}[ht]
\centering
\caption{Results for Meltome (human cell)}
\begin{tabular}{lcc cc}
\toprule
Model & \multicolumn{2}{c}{$\rho$} & \multicolumn{2}{c}{MSE} \\
\cmidrule(lr){2-3} \cmidrule(lr){4-5}
                            & Train            & Test          & Train              & Test               \\
\midrule
\textit{esm1v}\textsuperscript{\cite{flip}}                & 0.780            & 0.670          & 12.37             & 18.16    \\
esm1v (6 Layers)            & 0.805            & 0.670          & 10.85             & 18.17             \\
esm2 (6 Layers)             & 0.690            & 0.650          & 16.55             & 19.65             \\
esm2 (33 Layers)            & 0.765            & 0.670          & 12.93             & 18.37             \\
esm2 (48 Layers)            & \textbf{0.960}   & 0.610          & \textbf{2.435}    & 23.23            \\
SaProt (ESMFold unmasked) \footnote[3]  & 0.765           & \textbf{0.697} & 12.84           & \textbf{16.83}  \\
SaProt (ESMFold masked) \footnote[3]    & 0.742           & 0.685          & 14.26           & 17.34           \\
\bottomrule
\end{tabular}
\label{tab:meltome_human_cell}
\end{table}

\footnotetext[3]{Sequences longer than 2157 AA (162 samples) were masked completely due to computational constraints}

The evaluation of the models on the Meltome datasets reveals several interesting trends and provides insights into the trade-offs between training performance and generalization ability. In general, deeper models tend to achieve higher correlations and lower errors on the training data, but this often comes at the expense of lower generalization to the test set. This is evident in the ESM-2 model with 48 layers, which scores high correlation in the training set but does not hold this performance in the test set.

The observed differences between sequence-only models (ESM) and structure-aware models (SaProt) highlight the impact of incorporating structural information into predictions. Structure-aware models consistently exhibit improved generalization, suggesting that leveraging additional context beyond sequence embeddings contributes to capturing more robust patterns in protein fitness prediction tasks. Furthermore, the masking strategy employed in some models does not drastically alter performance in this split.

While the overall accuracy across models is relatively similar, the SaProt model stands out for its robust generalization, followed closely by the ESM-2 model with 33 layers. This suggests that balancing model complexity with architectural and training innovations is key to achieving strong and reliable performance across datasets.

\subsection{GB1}

%% \begin{table}[ht]
%%\centering
%%\caption{Results for GB1 (one vs rest)}
%%\begin{tabular}{lcc cc}
%%\toprule
%%Model & \multicolumn{2}{c}{$\rho$} & \multicolumn{2}{c}{MSE} \\
%%\cmidrule(lr){2-3} \cmidrule(lr){4-5}
%%                            & Train     & Test          & Train     & Test      \\
%%\midrule
%%esm1v (FLIP)                & 0.65      & 0.09          & 0.70      & 1.92      \\
%%esm1v                       & X         & X             & X         & X         \\
%%esm2 (6 Layers)             & X         & X             & X         & X         \\
%%esm2 (33 Layers)            & X         & X             & X         & X         \\
%%esm2 (48 Layers)            & X         & X             & X         & X         \\
%%SaProt (ESMFold unmasked)   & X         & X             & X         & X         \\
%%SaProt (ESMFold masked)     & X         & X             & X         & X         \\
%%\bottomrule
%%\end{tabular}
%%\label{tab:gb1_one_vs_rest}
%%\end{table}

\begin{table}[ht]
\centering
\caption{Results for GB1 (two vs rest)}
\begin{tabular}{lcc cc}
\toprule
Model & \multicolumn{2}{c}{$\rho$} & \multicolumn{2}{c}{MSE} \\
\cmidrule(lr){2-3} \cmidrule(lr){4-5}
                            & Train             & Test              & Train             & Test           \\
\midrule
\textit{esm1v}\textsuperscript{\cite{flip}}                & 0.650             & 0.370             & 0.480             & 1.480          \\
esm1v (6 Layers)            & 0.770             & 0.520             & 0.315             & 1.355          \\
esm2 (6 Layers)             & 0.460             & 0.465             & 0.750             & 1.365          \\
esm2 (33 Layers)            & 0.790             & 0.590             & 0.275             & 1.190          \\
esm2 (48 Layers)            & 0.880             & \textbf{0.635}    & 0.160             & 1.330          \\
SaProt (ESMFold unmasked)   & \textbf{0.922}    & 0.595             & \textbf{0.077}   & \textbf{1.067} \\
SaProt (ESMFold masked)     & 0.870             & 0.634             & 0.169             & 1.078          \\
\bottomrule
\end{tabular}
\label{tab:gb1_two_vs_rest}
\end{table}

\begin{table}[ht]
\centering
\caption{Results for GB1 (three vs rest)}
\begin{tabular}{lcc cc}
\toprule
Model & \multicolumn{2}{c}{$\rho$} & \multicolumn{2}{c}{MSE} \\
\cmidrule(lr){2-3} \cmidrule(lr){4-5}
                            & Train             & Test              & Train             & Test           \\
\midrule
\textit{esm1v}\textsuperscript{\cite{flip}}                 & 0.87              & 0.83              & 0.32              & 0.71           \\
esm1v (6 Layers)            & 0.865             & 0.835             & 0.235             & 0.650          \\
esm2 (6 Layers)             & 0.850             & 0.820             & 0.300             & 0.810          \\
esm2 (33 Layers)            & 0.905             & 0.840             & 0.185             & \textbf{0.540} \\
esm2 (48 Layers)            & 0.960             & \textbf{0.850}    & 0.055             & 0.565              \\
SaProt (ESMFold unmasked)   & 0.963             & 0.777             & 0.050             & 0.703          \\
SaProt (ESMFold masked)     & \textbf{0.966}    & 0.811             & \textbf{0.045}    & 0.738          \\
\bottomrule
\end{tabular}
\label{tab:gb1_three_vs_rest}
\end{table}

\begin{table}[ht]
\centering
\caption{Results for GB1 (low vs high)}
\begin{tabular}{lcc cc}
\toprule
Model & \multicolumn{2}{c}{$\rho$} & \multicolumn{2}{c}{MSE} \\
\cmidrule(lr){2-3} \cmidrule(lr){4-5}
                            & Train             & Test              & Train             & Test     \\
\midrule
\textit{esm1v}\textsuperscript{\cite{flip}}                 & 0.82              & 0.53              & 0.03              & 3.26     \\
esm1v (6 Layers)            & 0.82              & 0.540             & 0.02              & 3.075    \\
esm2 (6 Layers)             & 0.82              & 0.510             & 0.02              & 3.200    \\
esm2 (33 Layers)            & 0.82              & 0.535             & 0.02              & 3.065    \\
esm2 (48 Layers)            & \textbf{0.855}    & \textbf{0.545}    & \textbf{0.01}     & \textbf{2.920}\\
SaProt (ESMFold unmasked)   & 0.836             & 0.362             & 0.02              & 3.366    \\
SaProt (ESMFold masked)     & 0.842             & 0.408             & \textbf{0.01}     & 3.118    \\
\bottomrule
\end{tabular}
\label{tab:gb1_low_vs_high}
\end{table}

\begin{table}[ht]
\centering
\caption{Results for GB1 (sampled)}
\begin{tabular}{lcc cc}
\toprule
Model & \multicolumn{2}{c}{$\rho$} & \multicolumn{2}{c}{MSE} \\
\cmidrule(lr){2-3} \cmidrule(lr){4-5}
                            & Train             & Test              & Train             & Test      \\
\midrule
\textit{esm1v}\textsuperscript{\cite{flip}}                 & 0.940             & 0.920             & 0.17              & 0.22         \\
esm1v (6 Layers)            & 0.950             & 0.930             & 0.06              & 0.130         \\
esm2 (6 Layers)             & 0.940             & 0.920             & 0.11              & 0.185         \\
esm2 (33 Layers)            & 0.960             & \textbf{0.940}    & 0.04              & 0.120         \\
esm2 (48 Layers)            & 0.960             & \textbf{0.940}    & 0.04              & \textbf{0.115}         \\
SaProt (ESMFold unmasked)   & \textbf{0.970}    & 0.873             & \textbf{0.01}   & 0.313         \\
SaProt (ESMFold masked)     & 0.961             & 0.913             & 0.02            & 0.207         \\
\bottomrule
\end{tabular}
\label{tab:gb1_sampled}
\end{table}

In the GB1 split, a slight shift in trends can be observed compared to the meltome split. In this case, the ESM-2 model with 48 layers outperforms the others, demonstrating better generalization to the test set and achieving the best overall results. The SaProt model, while excelling in training and achieving the highest training scores, does not generalize as effectively as the larger ESM-2 model on the test set, regardless of whether the structures are masked or unmasked. Despite this, SaProt achieves the best performance in the ``one vs. rest'' split, suggesting its strengths in handling specific tasks. Notably, in this split, masking low-certainty structures improves the model's test set performance, highlighting the potential advantages of this approach in certain scenarios.

\subsection{AAV}

\begin{table}[ht]
\centering
\caption{Test Results for AAV (two vs many)}
\begin{tabular}{lcc cc}
\toprule
Model                 & \multicolumn{2}{c}{$\rho$} & \multicolumn{2}{c}{MSE} \\
\cmidrule(lr){2-3} \cmidrule(lr){4-5}
                            & Train             & Test              & Train             & Test      \\
\midrule
\textit{esm1v}\textsuperscript{\cite{flip}}                 & \textbf{0.870}    & \textbf{0.70}              & \textbf{1.82}              & \textbf{6.00}      \\  
esm1v (6 Layers)            & 0.410             & 0.090             & 12.87            & 16.9    \\
esm2 (6 Layers)             & 0.485             & 0.145             & 9.505             & 12.8    \\
esm2 (33 Layers)            & 0.810             & 0.590    & 2.120             & 7.94 \\
\bottomrule
\end{tabular}
\label{tab:aav_two_vs_many}
\end{table}

\begin{table}[ht]
\centering
\caption{Test Results for AAV (seven vs many)}
\begin{tabular}{lcc cc}
\toprule
Model                 & \multicolumn{2}{c}{$\rho$} & \multicolumn{2}{c}{MSE} \\
\cmidrule(lr){2-3} \cmidrule(lr){4-5}
                      & Train             & Test              & Train            & Test      \\
\midrule
\textit{esm1v}\textsuperscript{\cite{flip}}           & \textbf{0.91}              & \textbf{0.70}              & \textbf{2.22}             & \textbf{2.58}      \\  
esm1v (6 Layers)      & 0.815             & 0.485             & 4.50             & 6.83     \\
esm2 (6 Layers)       & 0.745             & 0.390             & 4.38             & 7.80     \\
esm2 (33 Layers)      & 0.850             & 0.595    & 3.99             & 3.99 \\
\bottomrule
\end{tabular}
\label{tab:aav_seven_vs_many}
\end{table}

\begin{table}[ht]
\centering
\caption{Test Results for AAV (low vs high)}
\begin{tabular}{lcc cc}
\toprule
Model                 & \multicolumn{2}{c}{$\rho$} & \multicolumn{2}{c}{MSE} \\
\cmidrule(lr){2-3} \cmidrule(lr){4-5}
                      & Train             & Test              & Train             & Test      \\
\midrule
\textit{esm1v}\textsuperscript{\cite{flip}}           & \textbf{0.780}              & 0.350              & \textbf{1.11}              & \textbf{10.83}     \\  
esm1v (6 Layers)      & 0.295             & 0.065             & 2.645             & 29.93     \\
esm2 (6 Layers)       & 0.520             & 0.205             & 1.940             & 21.92     \\
esm2 (33 Layers)      & 0.720             & \textbf{0.365}    & 1.300             & 15.20 \\
\bottomrule
\end{tabular}
\label{tab:aav_low_vs_high}
\end{table}

Across the different AAV splits, the trend towards a better-performing ESM-2 (33 layers) model is evident, showcasing its ability to generalize better than smaller models. The results further emphasize the importance of model complexity, where deeper models with more parameters may offer significant advantages in certain contexts. However, careful consideration must be given to avoid overfitting, as shown by the performance of the ESM1v model in certain splits.
The SaProt and ESM-2 model with 48 layers could not be evaluated on the AAV split due to the extensive time required to compute embeddings and structures for this task.

\subsubsection{Impact of data in training}

\begin{figure}[!ht]
\centering
\begin{minipage}{0.495\textwidth}
  \centering
  \includegraphics[width=\linewidth]{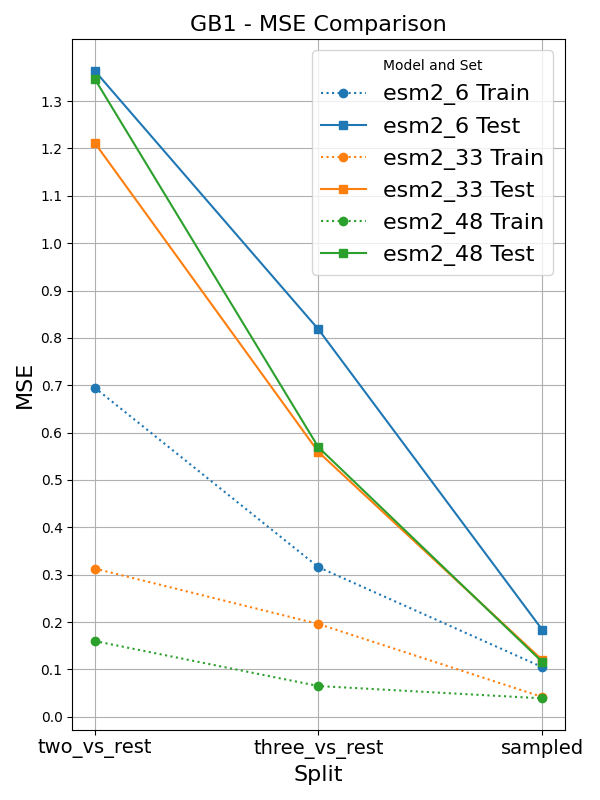} % Add image file name here
  \label{fig:mse}
\end{minipage}
\begin{minipage}{0.495\textwidth}
  \centering
  \includegraphics[width=\linewidth]{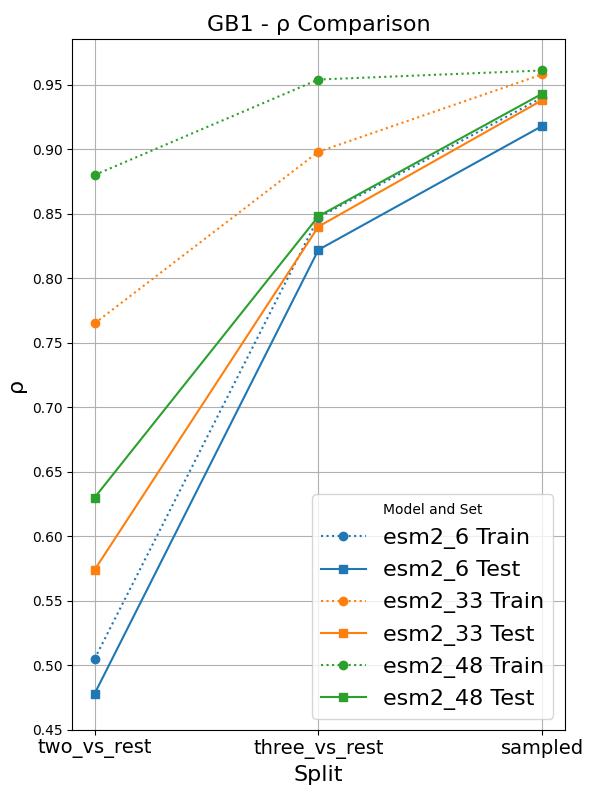} % Add image file name here
  \label{fig:rho}
\end{minipage}
\caption{Model comparison by split for both MSE and $\rho$ metrics (GB1)}
\label{fig:combined}
\end{figure}

To further assess how variations in the training data affect model performance, we focus on the results from the GB1 splits, where the splits clearly reflect the number of mutations in the training set. Figure \ref{fig:combined} illustrate how different splits in the training set impact performance. From left to right, each split includes progressively more data, as well as information from a higher number of mutations per sequence. It is evident that, with each richer split, performance improves for all models and the gap between training and evaluation narrows. However, from the plots we cannot clearly indicate whether the improvement is primarily due to the increased amount of data or the additional information about the number of mutations per sequence. A more detailed analysis is needed to determine with greater confidence whether a higher number of mutations per sequence in the training dataset contributes to improved evaluation across a broader range of mutations.
The results show that both the quantity of data and the diversity of mutations per sequence positively impact model performance, with richer training splits leading to improved results across all models.

\section{Conclusions and future work}
\label{conclusions}

In this study, we extended the FLIP benchmark by evaluating newer and larger models, including ESM-2 (in various sizes) and SaProt. By leveraging FLIP’s unique framework—training on low-mutation data and testing on high-mutation data—we assessed the ability of these models to generalize across different mutation landscapes. Our findings reveal that strong performance on low-mutation training and validation sets does not always translate to better generalization on high-mutation test sets. This highlights the critical need to prioritize generalization performance when applying these models in real-world scenarios.

Another important consideration in model selection is computational efficiency. While ESM-2 models were straightforward to implement, SaProt required a more complex pipeline due to its structure-aware design. Despite this, SaProt achieved superior correlation in the training set, whereas ESM-2 excelled in maintaining performance on the test set. These differences underscore the impact of training strategies and fine-tuning decisions. For example, reducing training duration or deprioritizing validation loss could mitigate overfitting and improve generalization to data with higher mutation rates.

Looking ahead, future research should focus on understanding how mutation density affects generalization loss in large protein models. By addressing this challenge, we can better tailor these models for practical applications, enhancing their robustness and effectiveness in real-world scenarios.

\newpage

\bibliographystyle{ieeetr}
\bibliography{bibliography}

\newpage

\section{Appendix}
\label{appendix}

% Loop through each split and metric
\begin{figure}[!ht]
    \centering
    % Example for one split
    \begin{subfigure}[b]{0.49\textwidth}
        \includegraphics[width=\textwidth]{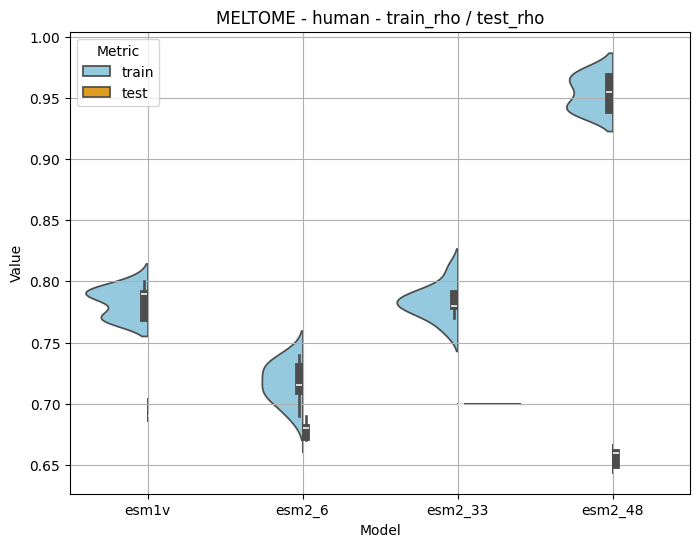}
        \caption{Human split: Train/Test Rho}
        \label{fig:violin_human_rho}
    \end{subfigure}
    \hfill
    \begin{subfigure}[b]{0.49\textwidth}
        \includegraphics[width=\textwidth]{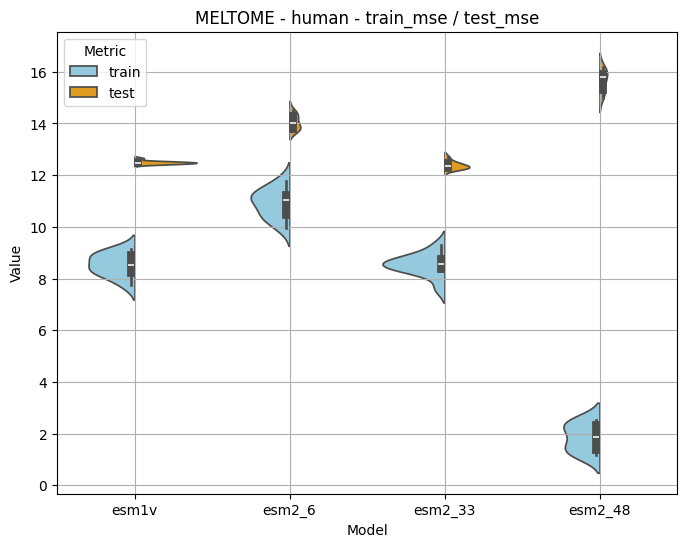}
        \caption{Human split: Train/Test MSE}
        \label{fig:violin_human_mse}
    \end{subfigure}

    \vspace{1em} % Adjust vertical spacing between rows

    \begin{subfigure}[b]{0.49\textwidth}
        \includegraphics[width=\textwidth]{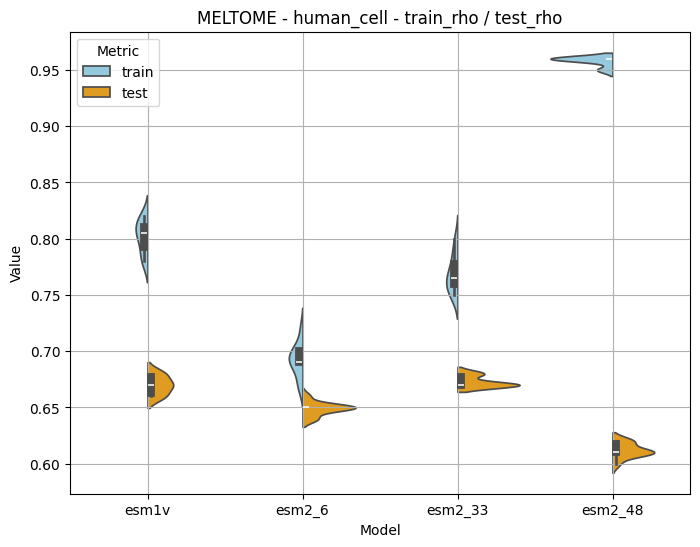}
        \caption{Human Cell split: Train/Test Rho}
        \label{fig:violin_human_cell_rho}
    \end{subfigure}
    \hfill
    \begin{subfigure}[b]{0.49\textwidth}
        \includegraphics[width=\textwidth]{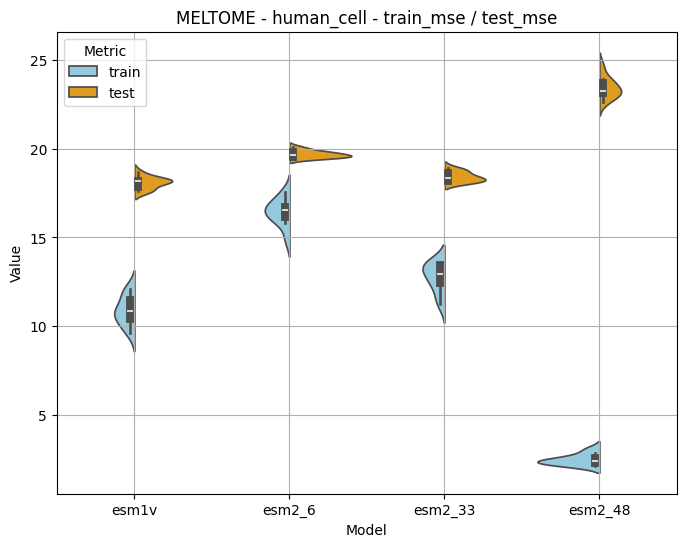}
        \caption{Human Cell split: Train/Test MSE}
        \label{fig:violin_human_cell_mse}
    \end{subfigure}

    \caption{Violin plots for metrics across splits and models for the Meltome dataset.}
    \label{fig:violin_plots_meltome}
\end{figure}

% Loop through each split and metric
\begin{figure}[!ht]
    \centering
    % Example for one split
    \begin{subfigure}[b]{0.49\textwidth}
        \includegraphics[width=\textwidth]{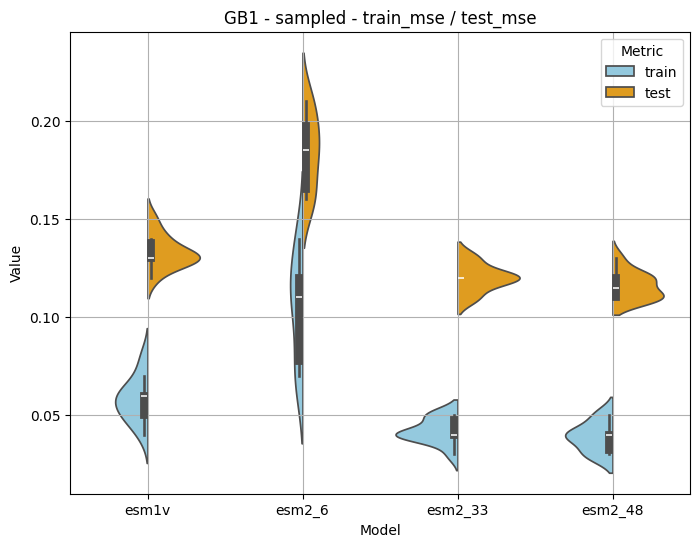}
        \caption{Sampled split: Train/Test MSE}
        \label{fig:violin_gb1_sampled_mse}
    \end{subfigure}
    \hfill
    \begin{subfigure}[b]{0.49\textwidth}
        \includegraphics[width=\textwidth]{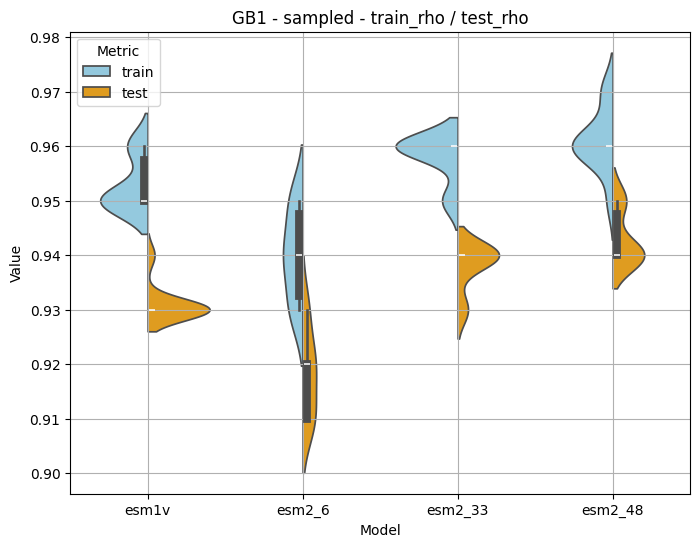}
        \caption{Sampled split: Train/Test Rho}
        \label{fig:violin_gb1_sampled_rho}
    \end{subfigure}

    \vspace{1em} % Adjust vertical spacing between rows

    \begin{subfigure}[b]{0.49\textwidth}
        \includegraphics[width=\textwidth]{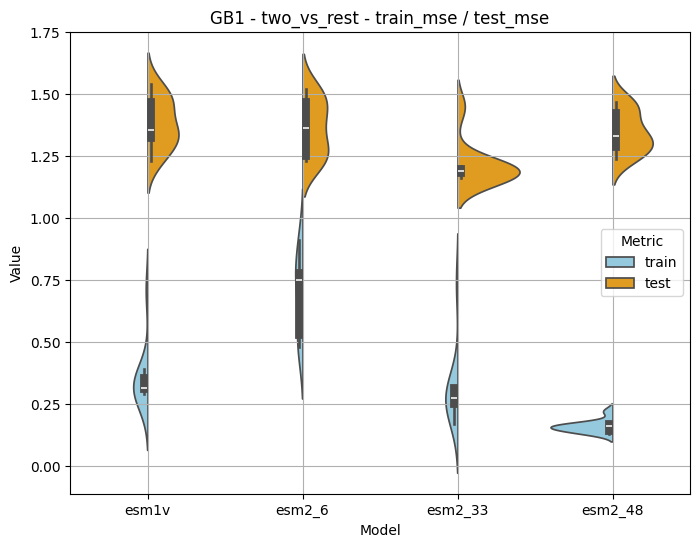}
        \caption{Two vs Rest split: Train/Test MSE}
        \label{fig:violin_gb1_two_mse}
    \end{subfigure}
    \hfill
    \begin{subfigure}[b]{0.49\textwidth}
        \includegraphics[width=\textwidth]{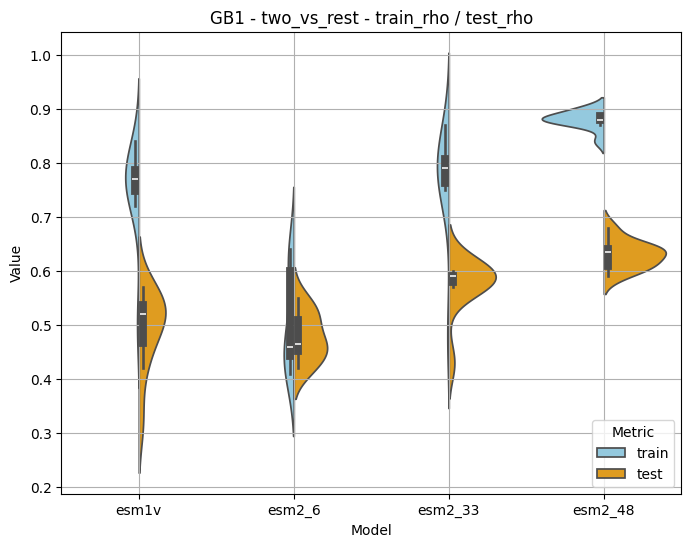}
        \caption{Two vs Rest split: Train/Test Rho}
        \label{fig:violin_gb1_two_rho}
    \end{subfigure}

    \vspace{1em} % Adjust vertical spacing between rows

    \begin{subfigure}[b]{0.49\textwidth}
        \includegraphics[width=\textwidth]{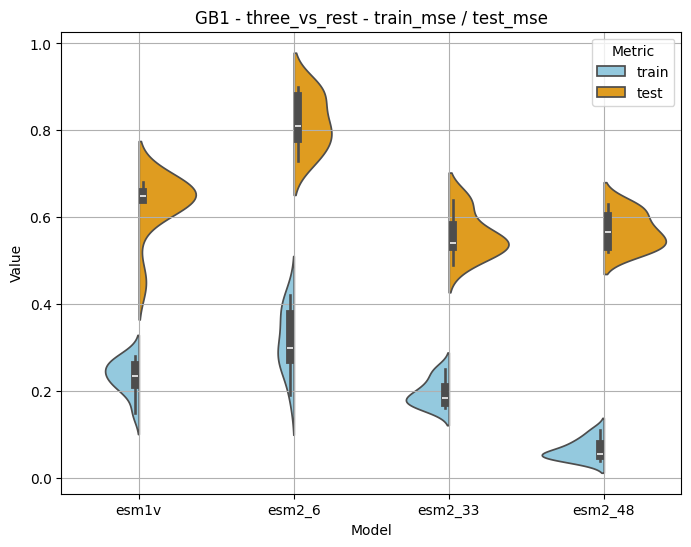}
        \caption{Three vs Rest split: Train/Test MSE}
        \label{fig:violin_gb1_three_mse}
    \end{subfigure}
    \hfill
    \begin{subfigure}[b]{0.49\textwidth}
        \includegraphics[width=\textwidth]{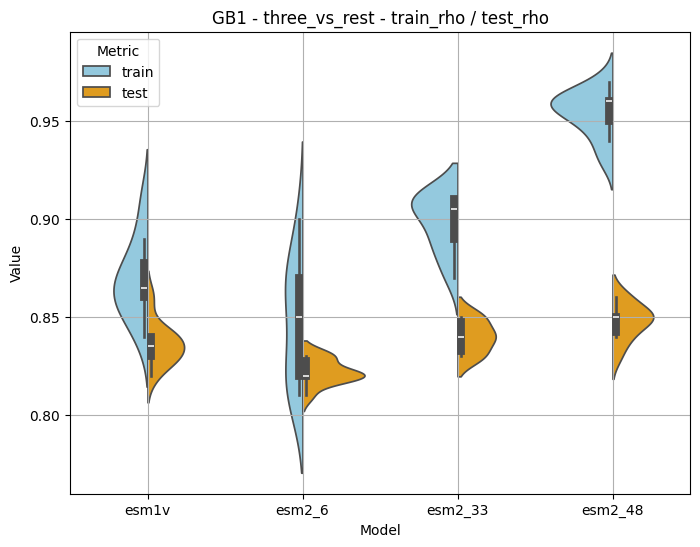}
        \caption{Three vs Rest split: Train/Test Rho}
        \label{fig:violin_gb1_three_rho}
    \end{subfigure}

    \caption{Violin plots for metrics across splits and models for the GB1 dataset.}
    \label{fig:violin_plots_gb1}
\end{figure}

% Loop through each split and metric
\begin{figure}[!ht]
    \centering
    % Example for one split
    \begin{subfigure}[b]{0.49\textwidth}
        \includegraphics[width=\textwidth]{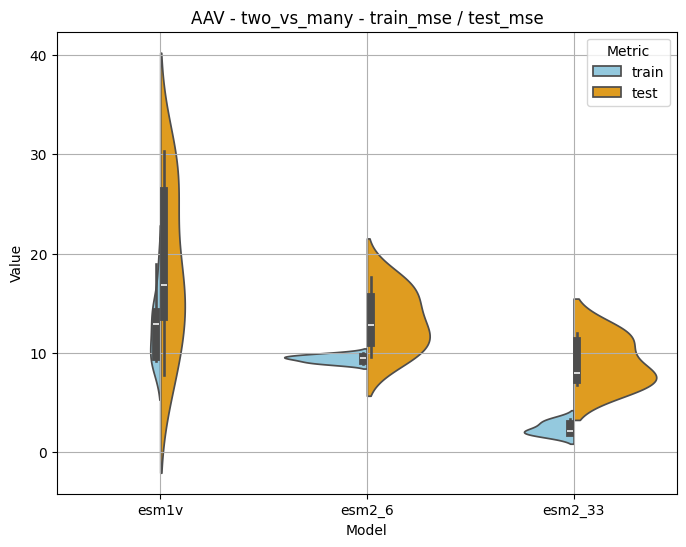}
        \caption{Two vs Many split: Train/Test MSE}
        \label{fig:violin_aav_two_mse}
    \end{subfigure}
    \hfill
    \begin{subfigure}[b]{0.49\textwidth}
        \includegraphics[width=\textwidth]{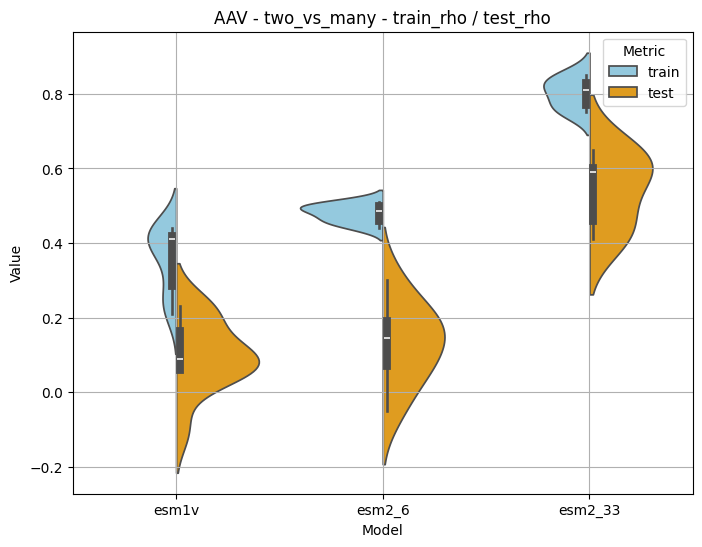}
        \caption{Two vs Many split: Train/Test Rho}
        \label{fig:violin_aav_two_rho}
    \end{subfigure}

    \vspace{1em} % Adjust vertical spacing between rows

    \begin{subfigure}[b]{0.49\textwidth}
        \includegraphics[width=\textwidth]{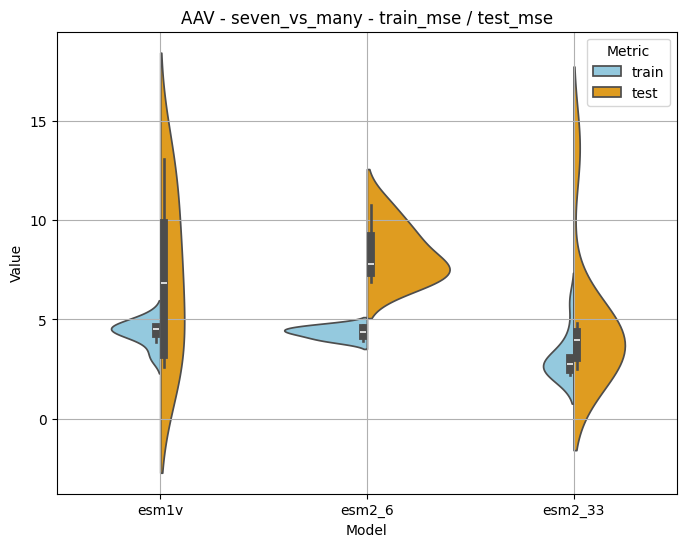}
        \caption{Seven vs Rest split: Train/Test MSE}
        \label{fig:violin_aav_seven_mse}
    \end{subfigure}
    \hfill
    \begin{subfigure}[b]{0.49\textwidth}
        \includegraphics[width=\textwidth]{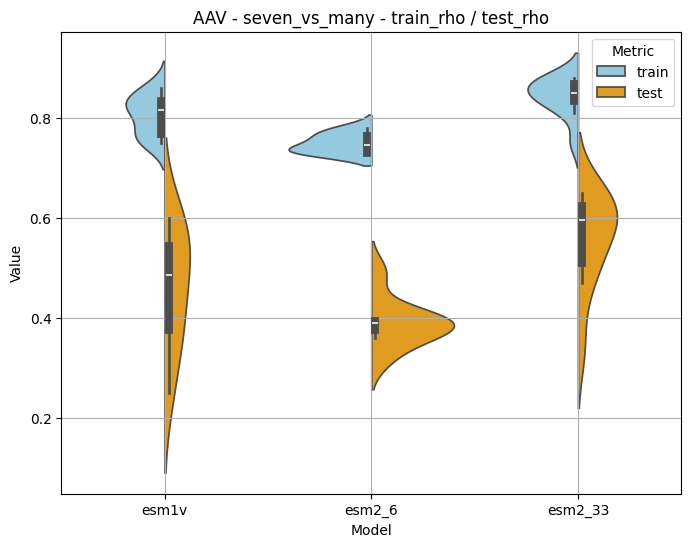}
        \caption{Seven vs Rest split: Train/Test Rho}
        \label{fig:violin_aav_seven_rho}
    \end{subfigure}

    \vspace{1em} % Adjust vertical spacing between rows

    \begin{subfigure}[b]{0.48\textwidth}
        \includegraphics[width=\textwidth]{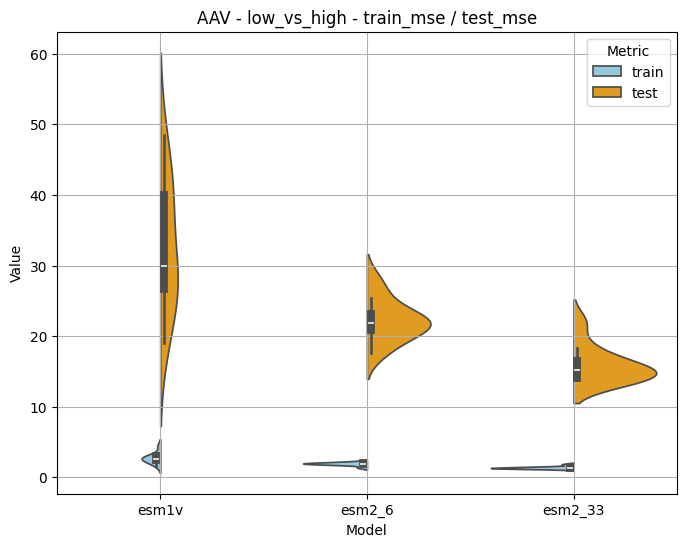}
        \caption{Low vs High split: Train/Test MSE}
        \label{fig:violin_aav_low_mse}
    \end{subfigure}
    \hfill
    \begin{subfigure}[b]{0.48\textwidth}
        \includegraphics[width=\textwidth]{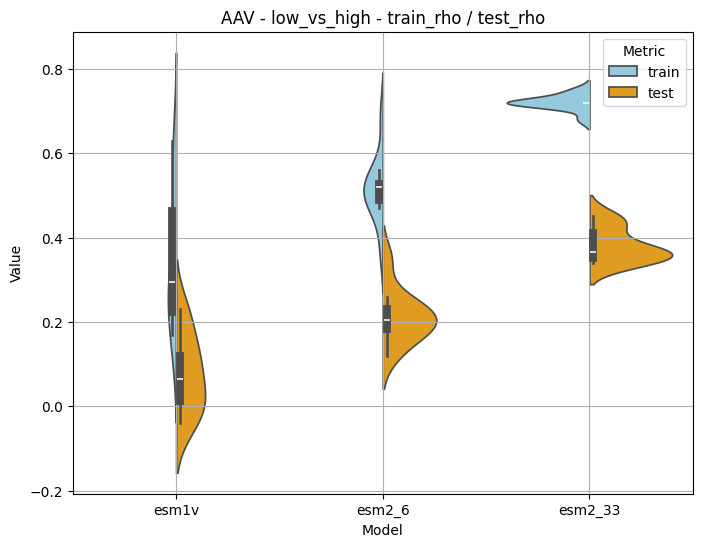}
        \caption{Low vs High split: Train/Test Rho}
        \label{fig:violin_aav_low_rho}
    \end{subfigure}

    \caption{Violin plots for metrics across splits and models for the AAV dataset.}
    \label{fig:violin_plots_aav}
\end{figure}

\end{document}